\def\MWEProblemsFourRevision{}
\documentclass[letterpaper]{article}
\usepackage[preprint]{aaai2027}
\usepackage[hyphens]{url}
\usepackage{graphicx}
\usepackage{natbib}
\usepackage{caption}
\usepackage{booktabs}
\usepackage{amsmath}

\usepackage{tabularx}
\usepackage{placeins}
\usepackage{xcolor}
\ifdefined\MWEReadingCopy
\usepackage[colorlinks=true,linkcolor=blue,citecolor=blue,urlcolor=blue]{hyperref}
\makeatletter
\let\mwe@orig@PackageError\PackageError
\def\PackageError#1#2#3{%
  \def\mwe@pkg{#1}%
  \def\mwe@aaai{aaai}%
  \ifx\mwe@pkg\mwe@aaai
    \PackageWarning{#1}{#2}%
  \else
    \mwe@orig@PackageError{#1}{#2}{#3}%
  \fi
}
\AtBeginDocument{\let\PackageError\mwe@orig@PackageError}
\makeatother
\fi
\newcommand{\wei}[1]{#1}
\newcommand{\aline}[1]{#1}
\ifdefined\MWEHighlightOriginalCorrections
\newcommand{\originalcorrection}[1]{\textcolor{blue}{#1}}
\else
\newcommand{\originalcorrection}[1]{#1}
\fi

\newcommand{\FailureRows}{%
cold feet & 19 & 19 & 0 \\
silver lining & 16 & 16 & 0 \\
throw in the towel & 10 & 10 & 0 \\
elephant in the room & 8 & 8 & 0 \\
green light & 8 & 8 & 0 \\
}

\newcommand{\FailurePairRows}{%
cold feet & 19 & silver lining & 16 \\
throw in the towel & 10 & elephant in the room & 8 \\
green light & 8 & & \\
}

\newcommand{\PromptFitMainRows}{%
Qwen3.6-35B-A3B & 256K & 0.940 & 1.000 & 0.480 \\
Gemma-4-26B-A4B-it & 256K & 0.880 & 1.000 & 0.500 \\
DeepSeek V4 Flash & 512K & 0.860 & 1.000 & 0.500 \\
DeepSeek V4 Flash & 1M & 0.880 & 1.000 & 0.480 \\
}

\newcommand{\AnchorPositionRows}{%
Qwen/Qwen3.6-35B-A3B & 32768 & late & 20 & 20 & 0.85 & 0.70 \\
Qwen/Qwen3.6-35B-A3B & 32768 & middle & 20 & 20 & 0.85 & 0.70 \\
Qwen/Qwen3.6-35B-A3B & 32768 & start & 20 & 20 & 0.95 & 0.90 \\
Qwen/Qwen3.6-35B-A3B & 65536 & late & 20 & 19 & 0.84 & 0.60 \\
Qwen/Qwen3.6-35B-A3B & 65536 & middle & 20 & 19 & 0.84 & 0.60 \\
Qwen/Qwen3.6-35B-A3B & 65536 & start & 20 & 20 & 0.85 & 0.70 \\
Qwen/Qwen3.6-35B-A3B & 8192 & late & 20 & 20 & 0.95 & 0.90 \\
Qwen/Qwen3.6-35B-A3B & 8192 & middle & 20 & 20 & 0.85 & 0.70 \\
Qwen/Qwen3.6-35B-A3B & 8192 & start & 20 & 20 & 0.95 & 0.90 \\
}

\newcommand{\PrimaryTransitionRows}{%
Qwen3.6-Plus & 174 & 0 & 176 & 0 \\
DeepSeek V4 Flash & 159 & 11 & 180 & 0 \\
DeepSeek V4 Pro & 141 & 34 & 171 & 4 \\
MiniMax-M3 & 164 & 16 & 168 & 2 \\
GLM-4.7 & 171 & 0 & 179 & 0 \\
GLM-5.1 & 174 & 0 & 176 & 0 \\
Qwen3.6-35B & 154 & 23 & 172 & 1 \\
Gemma-4-26B & 165 & 13 & 172 & 0 \\
}

\newcommand{\PrimaryPreservationRows}{%
Qwen3.6-Plus & 176 & 1.000 [1.000, 1.000] & 0.000 & 1.000 & 1.000 \\
DeepSeek V4 Flash & 180 & 1.000 [1.000, 1.000] & 0.000 & 0.937 & 0.937 \\
DeepSeek V4 Pro & 175 & 0.977 [0.956, 0.994] & 0.023 & 0.806 & 0.808 \\
MiniMax-M3 & 170 & 0.988 [0.970, 1.000] & 0.012 & 0.910 & 0.910 \\
GLM-4.7 & 179 & 1.000 [1.000, 1.000] & 0.000 & 1.000 & 1.000 \\
GLM-5.1 & 176 & 1.000 [1.000, 1.000] & 0.000 & 1.000 & 1.000 \\
Qwen3.6-35B & 173 & 0.994 [0.982, 1.000] & 0.006 & 0.872 & 0.876 \\
Gemma-4-26B & 172 & 1.000 [1.000, 1.000] & 0.000 & 0.926 & 0.926 \\
}

\newcommand{\SenseAsymmetryDetailedRows}{%
Qwen3.6-Plus & 1.000 & 1.000 & 0.000 & 1.000 (156) & 1.000 (18) & 1.000 & 1.000 & 1.000 \\
DeepSeek V4 Flash & 0.937 & 1.000 & -0.063 & 0.929 (156) & 1.000 (14) & 1.000 & 1.000 & 0.703 \\
DeepSeek V4 Pro & 0.823 & 0.960 & -0.137 & 0.797 (143) & 0.867 (30) & 0.967 & 0.979 & 0.379 \\
MiniMax-M3 & 0.920 & 0.977 & -0.057 & 0.911 (158) & 0.909 (22) & 1.000 & 0.987 & 0.703 \\
GLM-4.7 & 1.000 & 1.000 & 0.000 & 1.000 (154) & 1.000 (17) & 1.000 & 1.000 & 1.000 \\
GLM-5.1 & 1.000 & 1.000 & 0.000 & 1.000 (153) & 1.000 (21) & 1.000 & 1.000 & 1.000 \\
Qwen3.6-35B & 0.886 & 0.977 & -0.091 & 0.875 (160) & 0.875 (16) & 1.000 & 0.994 & 0.703 \\
Gemma-4-26B & 0.926 & 1.000 & -0.074 & 0.924 (170) & 1.000 (8) & 1.000 & 1.000 & 0.250 \\
}

\newcommand{\PriorStabilityEndpointRows}{%
Qwen3.6-Plus & 128K & 0.920 & 0.080 & 0.900 & 0.340 \\
DeepSeek V4 Flash & 128K & 0.920 & 0.080 & 0.920 & 0.320 \\
DeepSeek V4 Pro & 128K & 0.840 & 0.160 & 0.860 & 0.260 \\
MiniMax-M3 & 128K & 0.900 & 0.100 & 0.920 & 0.400 \\
GLM-4.7 & 128K & 0.820 & 0.180 & 0.980 & 0.380 \\
GLM-5.1 & 128K & 0.960 & 0.040 & 0.900 & 0.380 \\
Qwen3.6-35B & 128K & 0.940 & 0.060 & 0.920 & 0.400 \\
Gemma-4-26B & 128K & 0.860 & 0.140 & 1.000 & 0.400 \\
}

\newcommand{\RetentionEndpointRows}{%
Qwen3.6-Plus & 128K & 50 & 1.000 [1.000, 1.000] & 0.000 \\
DeepSeek V4 Flash & 128K & 50 & 0.940 [0.860, 1.000] & 0.060 \\
DeepSeek V4 Pro & 128K & 48 & 0.896 [0.812, 0.978] & 0.104 \\
MiniMax-M3 & 128K & 47 & 0.872 [0.783, 0.957] & 0.128 \\
GLM-4.7 & 128K & 50 & 1.000 [1.000, 1.000] & 0.000 \\
GLM-5.1 & 128K & 50 & 1.000 [1.000, 1.000] & 0.000 \\
Qwen3.6-35B & 128K & 48 & 0.938 [0.860, 1.000] & 0.062 \\
Gemma-4-26B & 128K & 50 & 0.900 [0.820, 0.980] & 0.100 \\
}

\newcommand{\ChineseBehavioralRows}{%
DeepSeek V4 Flash & 0.993 & 1.000 & 70 & 0.986 & 1.000 \\
DeepSeek V4 Pro & 0.864 & 0.936 & 70 & 0.843 & 0.886 \\
GLM-4.7 & 1.000 & 1.000 & 74 & 1.000 & 1.000 \\
GLM-5.1 & 1.000 & 1.000 & 69 & 1.000 & 1.000 \\
Qwen3.6-Plus & 1.000 & 1.000 & 72 & 1.000 & 1.000 \\
MiniMax-M3 & 0.886 & 0.921 & 70 & 0.771 & 1.000 \\
Qwen3.6-35B & 0.950 & 0.986 & 70 & 0.900 & 1.000 \\
Gemma-4-26B & 0.943 & 0.979 & 70 & 0.886 & 1.000 \\
}

\newcommand{\BilingualBehavioralRows}{%
Qwen3.6-Plus & 1.000 & 1.000 & 1.000 (174) & 1.000 (174) & 1.000 & 1.000 & 1.000 (72) & 1.000 \\
DeepSeek V4 Flash & 0.969 & 1.000 & 0.935 (170) & 0.935 (170) & 0.993 & 1.000 & 0.986 (70) & 1.000 \\
DeepSeek V4 Pro & 0.891 & 0.991 & 0.806 (175) & 0.809 (173) & 0.864 & 0.936 & 0.843 (70) & 0.886 \\
MiniMax-M3 & 0.949 & 0.991 & 0.911 (180) & 0.911 (180) & 0.886 & 0.921 & 0.771 (70) & 1.000 \\
GLM-4.7 & 1.000 & 1.000 & 1.000 (171) & 1.000 (171) & 1.000 & 1.000 & 1.000 (74) & 1.000 \\
GLM-5.1 & 1.000 & 1.000 & 1.000 (174) & 1.000 (174) & 1.000 & 1.000 & 1.000 (69) & 1.000 \\
Qwen3.6-35B & 0.931 & 0.997 & 0.870 (177) & 0.875 (176) & 0.950 & 0.986 & 0.900 (70) & 1.000 \\
Gemma-4-26B & 0.963 & 0.997 & 0.927 (178) & 0.927 (178) & 0.943 & 0.979 & 0.886 (70) & 1.000 \\
}

\newcommand{\SharedORRetRange}{0.809--1.000}
\newcommand{\DSProORRet}{0.809}
\newcommand{\DSProORRetDen}{173}
\newcommand{\DSProGapCI}{0.088,0.295}
\newcommand{\MiniMaxGapCI}{0.038,0.148}

\newcommand{\HardCuePairedPilotDeltaRows}{%
DeepSeek V4 Flash & 10 & 60 & 0.067 [0.000, 0.167] & -0.100 [-0.217, 0.000] & -0.350 [-0.500, -0.200] & 0.000 [0.000, 0.000] \\
GLM-5.1 & 10 & 60 & 0.033 [0.000, 0.083] & 0.033 [0.000, 0.083] & -0.167 [-0.317, -0.050] & 0.000 [0.000, 0.000] \\
}

\newcommand{\HardCuePairedAllFamilyDeltaRows}{%
DeepSeek V4 Pro & 25 & 100 & 0.060 [0.020, 0.100] & -0.080 [-0.140, -0.020] & -0.250 [-0.320, -0.180] & -0.140 [-0.200, -0.080] \\
Qwen3.6-Plus & 25 & 100 & 0.000 [-0.030, 0.030] & -0.060 [-0.130, -0.010] & -0.380 [-0.460, -0.300] & 0.000 [0.000, 0.000] \\
GLM-4.7 & 25 & 100 & 0.000 [0.000, 0.000] & 0.000 [0.000, 0.000] & 0.020 [-0.060, 0.090] & 0.000 [0.000, 0.000] \\
MiniMax-M3 & 25 & 100 & 0.050 [-0.010, 0.110] & 0.000 [-0.050, 0.050] & -0.110 [-0.200, -0.040] & -0.060 [-0.110, -0.020] \\
Qwen3.6-35B & 25 & 100 & 0.050 [0.000, 0.110] & 0.020 [-0.020, 0.060] & -0.330 [-0.410, -0.250] & 0.000 [0.000, 0.000] \\
Gemma-4-26B & 25 & 100 & 0.040 [-0.010, 0.100] & -0.010 [-0.050, 0.030] & -0.190 [-0.280, -0.100] & 0.000 [0.000, 0.000] \\
}

\newcommand{\ConflictRetrievalRange}{0.989--1.000}
\newcommand{\DSProConflictRetrieval}{0.989}
\newcommand{\DSProConflictOverride}{0.806}
\newcommand{\DSProWithinGap}{0.183}
\newcommand{\DSProWithinHolmP}{0.002}
\newcommand{\WithinModelDissociationRows}{%
Qwen3.6-Plus & 174 & 1.000 & 1.000 & 0.000 & 0 & 0 & 1.000 & 1.000 \\
DeepSeek V4 Flash & 170 & 1.000 & 0.935 & 0.065 & 11 & 0 & 0.125 & 0.500 \\
DeepSeek V4 Pro & 175 & 0.989 & 0.806 & 0.183 & 33 & 1 & $<0.001$ & 0.002 \\
MiniMax-M3 & 180 & 1.000 & 0.911 & 0.089 & 16 & 0 & 0.008 & 0.055 \\
GLM-4.7 & 171 & 1.000 & 1.000 & 0.000 & 0 & 0 & 1.000 & 1.000 \\
GLM-5.1 & 174 & 1.000 & 1.000 & 0.000 & 0 & 0 & 1.000 & 1.000 \\
Qwen3.6-35B & 177 & 0.994 & 0.870 & 0.124 & 22 & 0 & 0.008 & 0.055 \\
Gemma-4-26B & 178 & 1.000 & 0.927 & 0.073 & 13 & 0 & 0.031 & 0.156 \\
}

\newcommand{\HostedConsensusHardN}{182}

\newcommand{\HostedConsensusDSFlash}{0.940}
\newcommand{\HostedConsensusDSPro}{0.813}
\newcommand{\HostedConsensusMiniMax}{0.918}
\newcommand{\HostedConsensusGLMFourSeven}{1.000}

\newcommand{\ColdFeetMissCount}{19}

\newcommand{\SameCallDualHeadCompactRows}{%
DeepSeek V4 Pro & 0.920/1.000 & 0.920/1.000 & 0.920/1.000 & 0.900/1.000 \\
GLM-5.1 & 1.000/1.000 & 1.000/1.000 & 1.000/1.000 & 1.000/1.000 \\
MiniMax-M3 & 0.940/1.000 & 0.940/1.000 & 0.940/1.000 & 0.920/0.960 \\
}

\title{MWE-ECL: Recoverable Long-Range Context Does Not Always Override Local Lexical Priors}

\author{
Wei He\textsuperscript{1},
Aline Villavicencio\textsuperscript{1},
Rodrigo Wilkens\textsuperscript{1},
Zhenyun Deng\textsuperscript{2}
}
\affiliations{
\textsuperscript{1}University of Exeter\\
\textsuperscript{2}University of Cambridge
}

\begin{document}

\maketitle

\begin{abstract}
Long-context evaluations often test whether a model can recover distant
evidence, but recoverability does not guarantee behavioral influence.
\wei{We test the prediction that a distant discourse anchor can remain
explicitly recoverable yet fail to change the locally preferred reading of
a familiar multiword expression; such failures should concentrate when the
model's no-anchor default conflicts with the anchor, while prior-correct
decisions remain largely preserved.}
\wei{We introduce Multiword Expression Effective Context Length (MWE-ECL),
a bilingual diagnostic whose matched anchor-retrieval, no-anchor prior, and
interpretation prompts measure explicit recoverability, model-observed
defaults, and anchor-conditioned decisions, respectively.}
\wei{Across eight English deployment panels on a shared 0--128K grid,
retrieval-control accuracy on prior-conflict items is 0.989--1.000,
prior-conflict override spans 0.806--1.000
(0.809--1.000 after conditioning on correct retrieval), and preservation
of prior-correct decisions remains 0.977--1.000.}
\wei{A same-call control querying retrieval and interpretation in one prompt
reproduces the gap for DeepSeek V4 Pro (1.000 retrieval versus
0.900--0.920 interpretation), showing that separate invocations are not its
sole explanation; smaller or absent gaps in the other two models bound its
generality.}
\wei{For DeepSeek V4 Flash, separate prompt-fit tests retain perfect
retrieval with lower interpretation at 512K and 1M, while foil-consistent
cues shift the no-anchor prior far more than retrieval; cross-model cue
effects are heterogeneous.}
\wei{A separately reported 10-family Chinese subset shows similar
descriptive gaps, 
but imperfect retrieval for some models prevents an
integration-only attribution.}
\wei{MWE-ECL therefore evaluates whether explicitly recoverable distant
context changes a competing local semantic decision.}
\end{abstract}

\section{Introduction}

\wei{A model may correctly report a constraint stated more than 100,000 tokens
earlier yet make the same later decision it would have made without that
constraint. A million-token context window specifies nominal capacity, not a
guarantee of effective use. Prior work operationalizes effective context length
(ECL) as the largest tested context length at which task performance remains
satisfactory \citep{hsieh2024ruler,an2024effectivecontext}. Here, we focus on whether
recoverable distant evidence changes a decision when it conflicts with a strong
local expectation.}

\wei{This distinction matters for long-running assistants and agents, where a short
current turn may depend on a preference, policy exception, or state update
established far earlier.}

Retrieval suites reveal position-dependent access failures
\citep{liu2023lost,hsieh2024ruler}, while broader evaluations show that exact
recoverability need not ensure use
\citep{goldman2024really,yang2024dolce,du2025context,chen2026longbenchpro}.
Multiword Expression Effective Context Length (MWE-ECL) operationalizes this
distinction as lexical-semantic prior override. Familiar multiword expressions
(MWEs) invite conventional local readings, but a distant discourse
anchor can license the competing sense \citep{haagsma2020magpie,mi2025dice}.
\aline{The model must therefore assess whether to use a distant but accessible state to change a local decision.}
\wei{In our design, increasing neutral filler jointly increases the
total context size and the distance between the relevant anchor and the
decision point.}

MWEs provide a controlled version of this conflict. For example,
\emph{spill the beans} normally favors a figurative reading, whereas a distant
warehouse scene can license a literal one. The local trigger is identical across
senses, so lexical familiarity cannot solve the task; \aline{it is the anchor that alters} 
the
decision. This isolates prior override without claiming to reproduce every
component of an agent workflow.

\begin{figure*}[ht!]
\centering
\includegraphics[height=0.2\textheight, keepaspectratio,width=\textwidth]{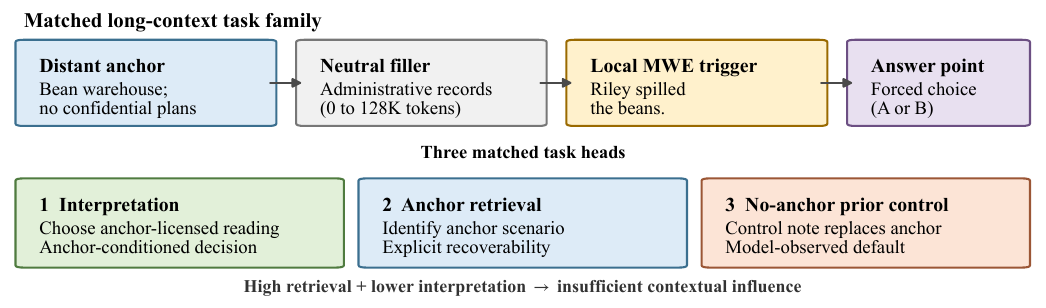}
\caption{MWE-ECL prompt timeline and matched controls. The heads separate
 (1) anchor-conditioned interpretation, (2) explicit recoverability, and (3) the
model-observed default; high retrieval with lower interpretation indicates
insufficient contextual influence at the behavioral level.}
\label{fig:overview}
\end{figure*}

\aline{Each literal/figurative case yields matched
interpretation, anchor-retrieval, and no-anchor prior-control prompts (Figure~\ref{fig:overview}).} 
These
heads separate explicit anchor access, the model-observed local default, and
the anchored decision. ECL is reported only as right-censored tested-range
coverage; prior-conflict override is the targeted context-use metric.

We ask three research questions. \textbf{RQ1 (Access):}
Can models identify the distant anchor when asked directly across the tested
distances? \textbf{RQ2 (Override):} When the anchor-free prompt elicits the
wrong interpretation, does the matched anchor-present prompt override that
choice? \textbf{RQ3 (Robustness):} Does the access--use gap persist at longer
distances, under semantic interference, in a second language, and within a
same-call control? 
This paper makes three contributions

\begin{itemize}
\item We introduce MWE-ECL, a bilingual long-context diagnostic whose matched
interpretation, retrieval-control, and no-anchor prior-control heads separate
explicit anchor recoverability, model-observed local defaults, and
anchor-conditioned decisions.
\item We define prior-conflict override and add retrieval-conditioned override
and preservation, distinguishing correction of a wrong local default from
access and corruption of an already correct default.
\item \wei{We reveal a recurring behavioral access--use dissociation in English across eight deployment panels and context scales, with descriptive supporting evidence under semantic interference and in Chinese.}
\end{itemize}

\section{Related Work}
\label{sec:related-work}

\paragraph{Long-context access and use.}
LongBench, RULER, NeedleBench, LongBench v2, and NoLiMa progressively broaden
long-context access, aggregation, reasoning, and low-overlap retrieval
\citep{bai2023longbench,hsieh2024ruler,li2024needlebench,bai2024longbenchv2,modarressi2025nolima}.
Efficient inference methods such as DuoAttention address the cost of
serving long contexts \citep{xiao2024duoattention}; our concern is
complementary: whether served context changes behavior.
These suites establish that nominal context length overstates effective access:
position effects remain substantial \citep{liu2023lost}, and retrieval becomes
harder as lexical overlap and evidence locality decrease. Yet a successful
lookup still leaves open whether the recovered state influences a downstream
choice. MWE-ECL therefore places an explicit retrieval control beside a
decision whose correct answer can oppose a strong local lexical default.

\paragraph{Beyond retrieval: integration and knowledge conflict.}
Beyond-retrieval benchmarks distinguish localized access from holistic or
dispersed-evidence integration
\citep{goldman2024really,yang2024dolce,du2025context,chen2026longbenchpro}.
Knowledge-conflict work likewise shows that accessible context can lose to
memorized associations \citep{xie2023adaptive,du2024contextprior,sun2026taskmatters,fu2026harnessing}.
These lines of work motivate a stricter question than fact recovery: whether
context changes a response when an alternative is already supported by
parametric or local evidence. MWE-ECL contributes a controlled lexical-semantic
version of this problem, rather than the general retrieval--use distinction
itself. Its no-anchor head measures the competing default on the same trigger,
and its retrieval head verifies that the distant scenario remains explicitly
recoverable without requiring the model to solve the semantic conflict.

\paragraph{Retrieval-Augmented Generation and Agent Memory.}
Retrieval-augmented generation (RAG) and memory evaluations include
retriever quality, faithfulness, temporal
reasoning, and updates \citep{hagstrom2025reality,wu2024longmemeval}. MWE-ECL
removes the external retriever: the anchor is already in context, eliminating
document selection as a source of failure. It also differs from answer-bearing
RAG evaluation because the anchor does not simply contain the final answer; it
licenses one reading of a later ambiguous expression. The test is therefore
whether available memory changes an activated interpretation, a pattern that
also arises when agents must apply an old preference or exception to a new turn.

\paragraph{Multiword expressions.}
MWEs are non-compositional and context-sensitive
\citep{sag2002mwe,baldwin2010mwe,constant2017survey}. MAGPIE, DICE, PaCE, and
MIDI support literal/idiomatic and multilingual analysis
\citep{haagsma2020magpie,mi2025dice,li2026pace,almheiri2026midi}; prior probes
are usually sentence- or short-context based and primarily assess sense
classification or contextual bias. Recent work investigates
idiomaticity in word representations \citep{he2025idiomaticity}; complementary
multimodal work uses semantic anchorage to quantify a semiotic gap
\citep{he2026semanticanchorage}. MWE-ECL instead separates a visible phrase
from its licensing evidence, pairs literal and figurative cases within each
family, and inserts controlled long-distance filler. Matched retrieval and
model-specific prior controls then distinguish failure to access the licensing
scenario from failure to use it.

\section{Benchmark Design}
\label{sec:method}

MWE-ECL is a diagnostic stress test, not \wei{a benchmark of average-case}
language
understanding. It intentionally selects familiar expressions 
\wei{for which a
conventional local reading competes with an alternative interpretation
licensed by distant discourse. Its matched interpretation,
anchor-retrieval, and no-anchor prior-control heads separate the
anchor-conditioned decision, explicit recoverability, and the
model-observed default. Jointly low retrieval and interpretation is
compatible with access failure; high retrieval with lower interpretation
indicates insufficient contextual influence; and low preservation
indicates anchor-induced corruption. These are behavioral, not
hidden-state, conclusions.}

\subsection{Task Structure}

Each anchor-present prompt contains a start-position \textbf{anchor}, deterministic neutral
\textbf{filler}, an end-position ambiguous \textbf{trigger}, and a binary
\textbf{question}. All eight primary panels use nominal
0/512/2K/8K/32K/64K/128K filler. K denotes 1,024 tokens.
Start placement and an explicit whole-document instruction favor access;
a small anchor-position pilot tests this placement choice.

The construction holds the local decision surface fixed while varying
the distant licensing context. Within each family, literal and figurative
variants share the same trigger sentence and answer options; the distant
anchor determines which interpretation is licensed, while the prior
control removes that anchor. Target-independent filler increases both
context size and anchor--decision distance without adding sense-relevant
evidence, so the primary grid does not disentangle their effects.
This design
trades document naturalism for a controlled contrast between access and use.

\begin{table}[t]
\centering
\footnotesize
\setlength{\tabcolsep}{3pt}
\renewcommand{\arraystretch}{1.04}
\ifdefined\MWEHighlightProblemsFour\color{blue}\fi
\begin{tabularx}{\columnwidth}{@{}>{\raggedright\arraybackslash\bfseries}p{0.25\columnwidth}>{\raggedright\arraybackslash}X@{}}
\toprule
Component & Matched content \\
\midrule
Literal condition & \textit{Anchor:} dry-goods warehouse that stores sacks of kidney beans, black beans, and lentils. The staff handles... \\
Figurative condition & \textit{Anchor:} team guarding confidential plans for a product launch. There are no legumes, groceries, kitchens, farms, or... \\
\midrule
Shared trigger & During the closing shift, Riley spilled the beans. \\
Option A & Riley revealed confidential information. \\
Option B & Riley caused actual beans to fall or scatter. \\
\midrule
Evaluation metadata & Reference answers (not shown to the model): literal condition B; figurative condition A. \\
\bottomrule
\end{tabularx}
\caption{\ifdefined\MWEHighlightProblemsFour\color{blue}\fi Matched zero-filler example. Opposing anchors reverse the reference answer for a shared trigger and option set; anchors are abbreviated, and longer records insert neutral filler.}
\label{tab:example-item}
\end{table}

Table~\ref{tab:example-item} illustrates a paired family: two distant
anchors license opposite readings of the same \emph{spill the beans}
trigger and option set. The forced-choice format eliminates the need for
a judge model; anchors are manually contrastive, and filler contains
neither the target MWE nor either licensing scenario.

Because provider tokenization differs, nominal 128K filler yields
approximately 136,671 and 153,552 prompt tokens on representative
DeepSeek and Qwen endpoints, respectively. We therefore use nominal
filler distance for cross-model display and retain exact prompt-token
counts in the reproducibility audit.

\subsection{Controls and Hard-Cue Extension}

The retrieval control asks for anchor-scenario identification rather
than MWE interpretation. The no-anchor prior control replaces the anchor
with a non-disambiguating note while retaining the trigger and answer
options, yielding an operational model default rather than a human
lexical prior. Because the main heads use separate calls, their
difference is a behavioral between-call contrast; an order-balanced
same-call control tests whether the retrieval--interpretation gap
persists within one invocation.

The hard-cue extension replaces neutral filler with target-consistent
cues supporting the anchor-licensed interpretation, foil-consistent
cues supporting its competitor, or mixed cues containing both, while
holding the anchor, trigger, and answer options fixed.

\subsection{Metrics}
\label{sec:metrics}

For model $m$, task $t$, and distance $d$, let $R_{m,t,d}$ denote the
matched case--repetition identities for which task $t$ yields a non-error
prediction. Accuracy is computed over these task-specific identities:
\begin{equation}
\mathrm{Acc}(m,t,d)
=
\frac{1}{|R_{m,t,d}|}
\sum_{r \in R_{m,t,d}}
\mathbf{1}\{\hat{y}_r = y_r\}.
\end{equation}
Here, $y_r$ is the gold label for identity $r$, and $\hat{y}_r$ is the
model prediction. At threshold $\tau=0.8$, joint ECL is the furthest
tested distance where both interpretation and retrieval pass. Because
distance cells are independently counterbalanced and non-monotone, ECL
is right-censored coverage, not an estimated failure boundary.
Confidence intervals (CIs) use 10,000 whole-family bootstrap resamples;
case-clustered estimates provide a sensitivity check.

\paragraph{Override rate.}
For the comparisons below, let
$R_{m,d}
=
R_{m,\mathrm{prior},d}
\cap
R_{m,\mathrm{int},d}$
denote the matched identities with non-error prior and interpretation
predictions. We define the \emph{override rate} on the model-specific
prior-conflict set:
\begin{equation}
\mathcal{C}_{m,d}
=
\{\,r \in R_{m,d} :
\hat{y}^{\mathrm{prior}}_r \neq y_r\,\}.
\end{equation}
\begin{equation}
\mathrm{OR}(m,d)
=
\frac{1}{|\mathcal{C}_{m,d}|}
\sum_{r \in \mathcal{C}_{m,d}}
\mathbf{1}\{\hat{y}^{\mathrm{int}}_r = y_r\}.
\end{equation}
Here, $\hat{y}^{\mathrm{prior}}_r$ and
$\hat{y}^{\mathrm{int}}_r$ denote the no-anchor prior-control and
interpretation predictions, respectively. OR is the fraction of
prior-conflict identities corrected once the anchor is present.

Aggregate OR pools its numerator and denominator over all tested
distances $D_m$ in model $m$'s panel:
\begin{equation}
\mathrm{OR}(m)
=
\frac{
\sum_{d \in D_m}
\sum_{r \in \mathcal{C}_{m,d}}
\mathbf{1}\{\hat{y}^{\mathrm{int}}_r = y_r\}
}{
\sum_{d \in D_m}
|\mathcal{C}_{m,d}|
}.
\end{equation}

To condition on retrieval success, let
$\mathcal{C}^{\mathrm{ret}}_{m,d}
=
\mathcal{C}_{m,d}
\cap
R_{m,\mathrm{ret},d}$
denote prior-conflict identities with a non-error retrieval prediction.
We report matched retrieval-conditioned override as
\begin{equation}
\mathrm{OR{\mid}Ret}(m,d)
=
\frac{
\sum_{r \in \mathcal{C}^{\mathrm{ret}}_{m,d}}
\mathbf{1}\{
\hat{y}^{\mathrm{ret}}_r = y_r
\land
\hat{y}^{\mathrm{int}}_r = y_r
\}
}{
\sum_{r \in \mathcal{C}^{\mathrm{ret}}_{m,d}}
\mathbf{1}\{
\hat{y}^{\mathrm{ret}}_r = y_r
\}
}.
\end{equation}
Here, $\hat{y}^{\mathrm{ret}}_r$ denotes the retrieval-control
prediction. OR$\mid$Ret conditions on success in a separate matched
call; it does not prove that the interpretation call retrieved the
anchor.

Correction is only one possible anchor effect. We therefore define
\begin{equation}
\begin{aligned}
\mathrm{Pres}(m,d)
&=
\frac{1}{|R_{m,d}\setminus\mathcal{C}_{m,d}|}
\sum_{r \in R_{m,d}\setminus\mathcal{C}_{m,d}}
\mathbf{1}\{\hat{y}^{\mathrm{int}}_r = y_r\},
\\
\mathrm{Corrup}(m,d)
&=
1-\mathrm{Pres}(m,d).
\end{aligned}
\end{equation}
The matched-valid set $R_{m,d}$ partitions into prior-conflict and
prior-correct identities; OR and preservation summarize interpretation
correctness on these two subsets, respectively. Corruption is the
complement of preservation, where the anchor changes an already-correct
default into an incorrect interpretation.

Aggregate OR$\mid$Ret and preservation entries likewise pool their
corresponding numerators and denominators over $d \in D_m$; they are not
unweighted averages of per-distance rates. Rates with zero denominators
are reported as NA, and every nonzero OR$\mid$Ret denominator is
reported.

These conditional rates are model-specific by design. Whenever
$\mathcal{C}_{m,d}$ is nonempty, a model that always follows its
no-anchor default has an override rate of zero regardless of overall
accuracy. For cross-model comparability, we additionally report
conflict-set overlap and results on a fixed consensus-hard subset in the
supplementary conflict-set tables. The supplementary hard-cue tables
also report integration under retrieval (IUR), defined as interpretation
accuracy conditioned on correct anchor retrieval over the same
identities.

\section{Dataset}
\label{sec:data}

The bilingual core comprises an English (EN) subset of 50 cases
over 25 MWE families and a Simplified Chinese (ZH) subset of
20 cases over 10 families, with one literal and one figurative
case per family (Table~\ref{tab:dataset_summary}). The English cases
span four construction types; for 38 English cases, the target expressions were cross-checked
against DICE \citep{mi2025dice} and MAGPIE
\citep{haagsma2020magpie}. All benchmark anchors,
triggers, questions, and options are newly authored.

\begin{table}[t]
\centering
\small
\setlength{\tabcolsep}{2.5pt}
\renewcommand{\arraystretch}{1.0}
\ifdefined\MWEHighlightPolish\color{blue}\fi
\begin{tabular*}{\columnwidth}{@{\extracolsep{\fill}}lrrrr@{}}
\toprule
Set & Fam. & Cases & Dist. & Rec./model \\
\midrule
English subset & 25 & 50 & 7 & 1,050 \\
Chinese subset & 10 & 20 & 7 & 420 \\
English expansion & 30 & 60 & 3 & 540 \\
\bottomrule
\end{tabular*}
\caption{Dataset composition. Each family supplies one literal and
one figurative case. Family mix is English (EN): 14 idiom, 8 compound, 2 prepositional,
1 verb-particle; Chinese (ZH): 4 idiom, 4 compound, 1 verb-object, 1 domain-derived. The
three-distance expansion is automatically checked, not human-validated or pooled.}
\label{tab:dataset_summary}
\end{table}

The idiom-heavy English catalog intentionally stresses cases with
strong figurative defaults, so results may therefore not generalize to constructions
with weaker or domain-specific priors. The separate 60-case English expansion
broadens coverage over 30 families at 0/8K/64K, but is not difficulty-matched,
human-validated, pooled, or used for ranking.

The Chinese subset selects 10 families from XMPIE
\citep{torunoglu2026xmpie} and retains only source expressions and sense
evidence; no source sentence is copied. It uses the same three tasks at seven
distances through 128K, but its 20 cases are reported separately because
difficulty is not matched across languages. Distances are approximate under
provider tokenization, and prompt-token counts are recorded when available.

Automated checks cover schema, class balance, provenance, prompt
construction, filler generation, and static leakage; filler contains
neither verbatim answer options nor target MWEs. Two English-proficient
annotators and two native-Chinese annotators, all blind to model
outputs, independently selected the anchor-licensed labels. They
matched all intended labels (100/100 English judgments and 40/40
Chinese judgments) and flagged no items.

\section{Experiments}
\label{sec:experiments}

\subsection{Models}

Eight application programming interface (API) deployment panels form
the headline comparison: DeepSeek V4 Flash/Pro
\citep{deepseekai2026deepseekv4}, MiniMax-M3
\citep{minimax2026m3}, Qwen3.6-Plus and Qwen3.6-35B-A3B
\citep{qwen2026qwen36plus,qwen36_35b_a3b}, GLM-4.7 and GLM-5.1
\citep{zhipu2026glm47,zhipu2026glm51}, and Gemma-4-26B-A4B-it
\citep{gemmateam2026gemma4}. All eight models complete the same
seven-distance 0--128K English grid (1,050 records/model) and the
corresponding Chinese grid (420 records/model), with the language
subsets reported separately. Comparisons across mismatched grids or
endpoint configurations are descriptive deployment snapshots, not
architecture rankings.

Separate stress regimes evaluate Qwen3.6-35B-A3B and
Gemma-4-26B-A4B-it near 256K, and DeepSeek V4 Flash at 512K and 1M.
Hard-cue tests include a 3,300-row DeepSeek V4 Flash study spanning
neutral, target-consistent, foil-consistent, and mixed conditions,
and all-family neutral/target/foil cue triads for six additional models
at 8K and 64K; matched pilot slices cover DeepSeek V4 Flash and GLM-5.1.

\subsection{Implementation}

Each chat-completion prompt instructs the model to use the entire
document and answer with one letter. Runs use temperature $0$ and a four-token answer
budget. The first explicit \texttt{A}/\texttt{B} is extracted; completed but
unparsed outputs are incorrect, whereas provider failures are excluded and
reported separately.

Five panels disable extended thinking; Qwen3.6-Plus and both GLMs use
provider defaults. 
Thus, failures occur under explicit
context-use-favoring instructions.
\subsection{Evaluation Protocol}

All ECL calculations use $\tau=0.8$, and confidence intervals use the
stated whole-family bootstrap procedure. Gold
A/B labels are balanced within every task--distance cell. Counterbalancing may
swap a case's option order across distances, so cells are coverage snapshots,
not longitudinal trajectories. Matched-grid, prompt-fit, hard-cue, and Chinese
evaluations remain separate whenever grids, filler semantics, or languages differ.

Repeated-run checks cover Qwen3.6-35B-A3B and
Gemma-4-26B-A4B-it at 64K and in three 128K passes; same-call controls
cover DeepSeek V4 Pro, GLM-5.1, and MiniMax-M3 with retrieval and
interpretation heads in the same call at 64K/128K. Other headline panels are
single-pass; bootstrap intervals quantify item/family variation, not API
nondeterminism. 

Primary within-model dissociation tests compare retrieval-control
accuracy and override on identical conflict rows. We use exact
family-clustered sign-flip tests with Holm correction over eight
pre-specified within-model tests; effect sizes and whole-family bootstrap
CIs remain primary. Cross-model tests are secondary because
model-specific conflict sets, grids, and endpoint configurations can differ.

The shared 0--128K EN grid is the primary evaluation. Same-call,
hard-cue, prompt-fit, expansion, and ZH analyses probe invocation
separation, semantic interference, longer-context stress, lexical
coverage, and cross-language robustness, respectively; these regimes
are not pooled into one ranking.

\section{Results}
\label{sec:results}

\begin{figure*}[t]
\centering
\begin{minipage}[c]{0.49\textwidth}
\centering
\includegraphics[height=0.25\textheight, keepaspectratio,width=\linewidth]{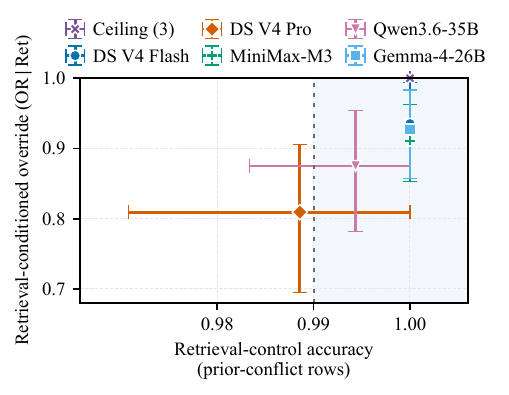}
\end{minipage}\hfill
\begin{minipage}[c]{0.49\textwidth}
\centering
\includegraphics[height=0.25\textheight,width=\linewidth]{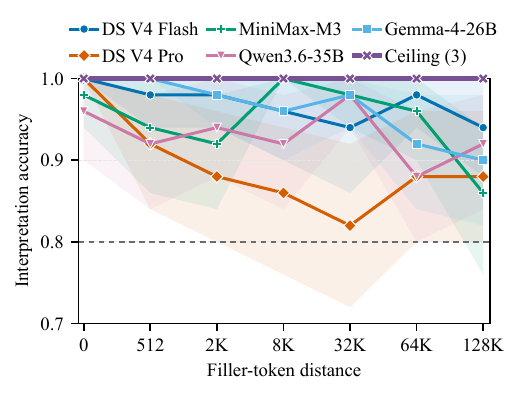}
\end{minipage}
\caption{Core English 0--128K results. \textbf{Left:} retrieval
versus OR$\mid$Ret on prior-conflict rows. \textbf{Right:} interpretation by
distance. Three models share the ceiling; bars/bands are family-bootstrap 95\%
CIs and cells are counterbalanced snapshots.}
\label{fig:retrieval-override}
\end{figure*}

\subsection{Near-Ceiling Recoverability Does Not Ensure Prior-Conflict Correction}

\textbf{RQ1--RQ2.} On shared-grid prior-conflict rows, retrieval is
\ConflictRetrievalRange{}, whereas OR and OR$\mid$Ret span 0.806--1.000 and
\SharedORRetRange{} (Figure~\ref{fig:retrieval-override}). DeepSeek V4
Pro retrieves at \DSProConflictRetrieval{} but attains
\DSProConflictOverride{} OR over 175 conflicts (gap \DSProWithinGap{}, 95\%
family-bootstrap CI $[\DSProGapCI]$; exact family sign-flip $p<0.001$, Holm
$p=\DSProWithinHolmP$). Its OR$\mid$Ret is \DSProORRet{}
($n=\DSProORRetDen{}$). MiniMax-M3's gap is 0.089 ($[\MiniMaxGapCI]$; Holm
$p=0.055$; conflict $n=180$). Thus recoverable evidence does not always change
the locally preferred interpretation.

\begin{table*}[t]
\centering
\scriptsize
\setlength{\tabcolsep}{2.6pt}
\renewcommand{\arraystretch}{0.96}
\ifdefined\MWEHighlightPolish\color{blue}\fi
\begin{tabular*}{\textwidth}{@{\extracolsep{\fill}}lrrrrrrrr@{}}
\toprule
& \multicolumn{4}{c}{English subset} &
  \multicolumn{4}{c}{Chinese subset} \\
\cmidrule(lr){2-5}\cmidrule(lr){6-9}
Model & Int. & Ret. & OR ($C$) & OR$\mid$Ret ($n$) &
Int. & Ret. & OR ($C$) & Pres. \\
\midrule
\BilingualBehavioralRows
\bottomrule
\end{tabular*}
\caption{English and Chinese \wei{subset} results (1,050/420 records per model), reported separately. $C$ and $n$ denote the OR and OR$\mid$Ret denominators; Pres. is prior-correct preservation. The subsets are not pooled or difficulty-matched.}
\label{tab:main}
\end{table*}

Table~\ref{tab:main} shows near-ceiling retrieval but lower
interpretation and override in several panels. 
\wei{All eight panels pass at the 128K endpoint under $\tau=0.8$, so
joint ECL is right-censored rather than a model ranking. At
$\tau=0.9$, joint ECL drops to 512 tokens for DeepSeek V4 Pro and
64K for MiniMax-M3; the other six panels remain right-censored at
128K.}

\wei{Aggregate prior-control accuracy against the anchor-licensed labels is
0.486--0.514, while semantic agreement between 0K and 128K no-anchor
choices, after mapping A/B to literal/figurative senses, is
0.820--0.960. Thus, the near-0.5 aggregate does not imply random or
wholly unstable defaults.}

At 128K, interpretation is 0.940/0.880/0.860 for DeepSeek V4
Flash/Pro and MiniMax-M3; Qwen3.6-35B/Gemma-4-26B reach 0.940/0.900, and
Qwen3.6-Plus, GLM-4.7, and GLM-5.1 remain at 1.000.

Balanced labels and same-order checks reduce simple option-position
explanations: on 146 shared 0--64K rows, DeepSeek V4 Pro still reaches 0.993
retrieval but 0.890 interpretation and 0.768 override (69 conflicts).

\subsection{The Gap Is Not Solely a Separate-Call Artifact}

\textbf{RQ3.} A targeted same-call control removes invocation separation by asking
both heads within one shared prompt. Each model completes 200 rows over 50
cases, two distances, and two counterbalanced query orders
(Table~\ref{tab:same-call-main}).

\begin{table}[t]
\centering
\scriptsize
\setlength{\tabcolsep}{2.4pt}
\ifdefined\MWEHighlightProblemsFour\color{blue}\fi
\begin{tabularx}{\columnwidth}{@{}>{\raggedright\arraybackslash}Xccc@{}}
\toprule
Model & Retrieval & Interpretation & Gap \\
\midrule
DeepSeek V4 Pro & 1.000 & 0.900--0.920 & 0.080--0.100 \\
MiniMax-M3 & 0.960--1.000 & 0.920--0.940 & 0.040--0.060 \\
GLM-5.1 & 1.000 & 1.000 & 0.000 \\
\bottomrule
\end{tabularx}
\caption{
\wei{Same-call dual-head control. Ranges are minima and maxima
across the four distance-by-query-order cells; Gap is retrieval minus
interpretation. Because the prior head requires an anchor-free prompt,
this control does not estimate override rate.}}
\label{tab:same-call-main}
\end{table}

Order effects are small and their family-bootstrap CIs include zero.
DeepSeek V4 Pro nevertheless makes four or five interpretation errors in each
50-case cell while retrieving all 50 anchors correctly. Separate invocations
therefore cannot be the sole source of its gap; the GLM-5.1 ceiling result also
bounds the claim's generality.

\subsection{Failures Concentrate Where Local Defaults Conflict}

\textbf{RQ2.} Because the suite is sense-balanced, aggregate
prior-control accuracy near 0.5 can hide a directional figurative
default. OR therefore conditions on each model's own prior errors.
Preservation is 0.977--1.000, indicating that anchors mostly correct
prior conflicts rather than disrupt already-correct defaults:
DeepSeek V4 Pro corrupts 4 of 175 prior-correct rows, MiniMax-M3
2 of 170, and Qwen3.6-35B 1 of 173; the other panels corrupt none.

Shared-row and fixed-conflict analyses show the same pattern. On
the consensus-hard 0--128K primary subset
($n{=}\HostedConsensusHardN{}$), \originalcorrection{fixed-subset interpretation
accuracies are}
\HostedConsensusDSFlash{}/\HostedConsensusDSPro{}, versus
\HostedConsensusMiniMax{}/\HostedConsensusGLMFourSeven{} for
MiniMax-M3/GLM-4.7. Literal-minus-figurative interpretation gaps are
$-0.063$ for DeepSeek V4 Flash, $-0.137$ for Pro, $-0.091$ for
Qwen3.6-35B, and $-0.074$ for Gemma. The corresponding
family-bootstrap CIs are $[-0.143,-0.006]$, $[-0.269,-0.017]$,
$[-0.194,0.011]$, and $[-0.143,-0.017]$, respectively; they exclude
zero for Flash, Pro, and Gemma, but not Qwen. Holm-adjusted sign-flip
tests remain descriptive ($p\geq0.250$), so we claim directional
concentration rather than a universal sense effect. OR$\mid$Ret sense
effects are heterogeneous.

All errors in the five highest-miss families occur on
literal variants (Table~\ref{tab:family-misses}).
\originalcorrection{For all \ColdFeetMissCount{} \emph{cold feet}
interpretation-error events, the matched retrieval prediction identifies the
literal anchor, whereas both the no-anchor prior and anchored interpretation
select the figurative option. This provides behavioral evidence that correct
matched retrieval can coexist with a response aligned with the local default.}

\ifdefined\MWEProblemsFourRevision
\begin{table}[t]
\centering
\scriptsize
\setlength{\tabcolsep}{3pt}
\renewcommand{\arraystretch}{0.90}
\ifdefined\MWEHighlightPresubmissionVFive\color{blue}\fi
\begin{tabularx}{\columnwidth}{@{}>{\raggedright\arraybackslash}Xrrr@{}}
\toprule
Family & Misses & Lit. & Fig. \\
\midrule
\FailureRows
\bottomrule
\end{tabularx}
\caption{English families with the most interpretation errors across
the eight 0--128K panels. Counts aggregate model--distance records; all families
tied at the cutoff are shown.}
\label{tab:family-misses}
\end{table}
\fi

\ifdefined\MWEProblemsFourRevision
\else
\begin{table*}[t]
\centering
\scriptsize
\setlength{\tabcolsep}{3pt}
\begin{minipage}[t]{0.46\textwidth}
\centering
\textit{(a) Highest-miss English families}\par\vspace{2pt}
\begin{tabular*}{\linewidth}{@{\extracolsep{\fill}}lrlr@{}}
\toprule
Family & N & Family & N \\
\midrule
\FailurePairRows
\bottomrule
\end{tabular*}
\end{minipage}\hfill
\begin{minipage}[t]{0.51\textwidth}
\centering
\textit{(b) Automatically checked catalog expansion}\par\vspace{2pt}
\begin{tabular}{lrrrr}
\toprule
Model & Int. & Ret. & OR & C \\
\midrule
DeepSeek V4 Pro & 0.944 & 0.961 & 0.897 & 87 \\
GLM-4.7 & 0.933 & 0.961 & 0.881 & 84 \\
GLM-5.1 & 0.922 & 0.983 & 0.878 & 90 \\
MiniMax-M3 & 0.928 & 0.983 & 0.867 & 90 \\
\bottomrule
\end{tabular}
\end{minipage}
\caption{Auxiliary robustness summaries. Left: all displayed misses
are literal variants. Right: 60 expansion cases at 0/8K/64K; this catalog is
not human-validated or difficulty-matched and therefore does not support
ranking. Full intervals and rows are supplementary.}
\label{tab:auxiliary-results}
\end{table*}
\fi

\subsection{Stress Tests Reproduce and Bound the Dissociation}

\textbf{RQ3.} Separate prompt-fit snapshots retain 1.000 retrieval
while interpretation remains lower at the 256K, 512K, and 1M settings
(Table~\ref{tab:prompt-fit-main}). Because these runs use
endpoint-specific prompt-fit configurations, they are not a monotone
extension of the matched grid.

\begin{table}[t]
\centering
\scriptsize
\setlength{\tabcolsep}{2pt}
\begin{tabularx}{\columnwidth}{@{}>{\raggedright\arraybackslash}Xcrrr@{}}
\toprule
Model & Context & Int. & Ret. & Prior \\
\midrule
\PromptFitMainRows
\bottomrule
\end{tabularx}
\caption{Prompt-fit stress outside the matched grid. Retrieval stays at ceiling
while interpretation is lower\ifdefined\MWEProblemsFourRevision.\else; full
accounting is supplementary.\fi}
\label{tab:prompt-fit-main}
\end{table}

With 4 and 16 foil-consistent cues, DeepSeek V4 Flash
interpretation falls to 0.873 and 0.867, respectively. At 16 cues, the no-anchor prior shifts far more than either retrieval or integration under retrieval (Table~\ref{tab:hard-cue-main}), supporting semantic
interference rather than simple anchor loss.

\begin{table}[t]
\centering
\small
\setlength{\tabcolsep}{2.0pt}
\renewcommand{\arraystretch}{0.95}
\ifdefined\MWEHighlightProblemsFour\color{blue}\fi
\ifdefined\MWEHighlightPresubmissionVFive\color{blue}\fi
\begin{tabular*}{\columnwidth}{@{\extracolsep{\fill}}lcc@{}}
\toprule
Metric & DS V4 Flash & Six-model min--max \\
\midrule
Outcome metric & IUR & Interpretation \\
$\Delta$Outcome & $-0.099$ & \originalcorrection{$-0.080$ to $0.020$} \\
$\Delta$Prior & $-0.378$ & \originalcorrection{$-0.380$ to $0.020$} \\
$\Delta$Retrieval & $-0.007$ & \originalcorrection{$-0.140$ to $0.000$} \\
\bottomrule
\end{tabular*}
\caption{Hard-cue outcomes at 16 foil cues. \originalcorrection{DeepSeek V4
Flash uses IUR across all 25 families; the six-model panels use interpretation
accuracy. The final column reports across-panel min--max ranges, not confidence
intervals. Deltas are foil minus neutral.}}
\label{tab:hard-cue-main}
\end{table}

Cue effects are heterogeneous. In the matched pilot, foil cues
change interpretation by $-0.100$ for DeepSeek V4 Flash and
$+0.033$ for GLM-5.1. In the all-family runs, the corresponding
changes are $-0.080$ for DeepSeek V4 Pro, $-0.060$ for
Qwen3.6-Plus, and $-0.010$ to $+0.020$ for the remaining four
models. Foil cues shift prior-control behavior more consistently
than interpretation or retrieval, whose effects remain
deployment-dependent.

The 60-case expansion also shows lower override than retrieval across
four models (override 0.867--0.897 vs.\ retrieval 0.961--0.983).
Because it is neither human-validated nor difficulty-matched, it serves
as a lexical-coverage check rather than ranking evidence.

The Chinese subset shows the same qualified pattern
(Table~\ref{tab:main}): three models are at joint ceiling, while
DeepSeek V4 Flash reaches 1.000 retrieval/0.986 OR and the other four non-ceiling
panels reach 0.921--0.986 retrieval/0.771--0.900 OR. At 128K, Qwen3.6-35B and
Gemma score 0.950/0.950 and 0.900/0.950 interpretation/retrieval. Because
retrieval itself is imperfect for some models and the subset has only 10
families, these results are descriptive, non-pooled, and not a cross-language
difficulty comparison. Preservation is 1.000 in seven panels, whereas
DeepSeek V4 Pro flips 8 of 70 prior-correct rows (Pres. 0.886); its Chinese gap
therefore combines weaker correction with anchor-induced corruption.

\ifdefined\MWEProblemsFourRevision
A 20-item Qwen3.6-35B-A3B position pilot holds the trigger at the end.
At 32K, start yields IUR/OR of 0.95/0.90 versus 0.85/0.70 at middle and late
positions. At 64K, the corresponding values are 0.85/0.70 versus 0.84/0.60;
each non-start IUR estimate conditions on 19 retrieval-correct rows. At 8K,
start and late yield 0.95/0.90 versus 0.85/0.70 at middle. The small-$N$ pilot
therefore does not rule out a start-position advantage.
\else
A supplementary 20-item Qwen3.6-35B-A3B anchor-position pilot moves
the anchor among start, middle, and late positions at 8K/32K/64K while holding
the trigger at the end; its small-$N$ design does not support a general
position-effect claim.
\fi

\ifdefined\MWEProblemsFourRevision
\else
\subsection{Same-Call Evidence Rules Out Invocation Separation Alone}

Signal loss alone cannot explain conflict-set retrieval of
0.989--1.000 alongside override of 0.806--1.000, nor perfect retrieval at the
512K/1M prompt-fit settings. The same-call control removes invocation separation
for a targeted three-model check: each model completes 200 rows over 50 cases,
two distances, and two counterbalanced query orders. DeepSeek V4 Pro retains
1.000 retrieval but 0.900--0.920 interpretation (gap 0.080--0.100);
MiniMax-M3 reaches 0.960--1.000 retrieval and 0.920--0.940 interpretation
(gap 0.040--0.060); GLM-5.1 is at 1.000 on both heads. Paired order effects are
small and their family-bootstrap CIs include zero. Thus separate invocations
cannot be the sole source of DeepSeek V4 Pro's gap: each 50-case cell contains
four or five interpretation errors despite all 50 retrieval decisions being
correct. Because the prior head requires anchor removal, it cannot be
included in the anchor-present call and this control does not estimate OR; the
ceiling GLM result bounds the claim's generality.
\fi

The evidence instead supports competition from a local lexical
default: literal errors concentrate in conventional idioms, hard cues shift the
prior more than retrieval, and \emph{cold feet} rows retrieve the literal anchor
while still selecting the figurative interpretation. This parallels contextual
knowledge conflict \citep{xie2023adaptive,fu2026harnessing}, but forced-choice
behavior does not identify a parameter-level mechanism.

\section{Discussion}

Near-ceiling retrieval is a control outcome, not evidence that the
diagnostic is uninformative. MWE-ECL intentionally makes the distant scenario
easy to identify under an explicit question, then asks whether the same evidence
changes a decision opposed by a familiar lexical default. The English ceiling
models show that the benchmark does not mechanically force an access--use gap;
the lower OR$\mid$Ret values of other panels provide the discriminating signal.
Likewise, perfect performance through 128K does not establish a model's context limit:
it only right-censors joint ECL at the largest tested endpoint.

Override answers a conditional question---whether context repairs a
model's own wrong default---and its denominator is consequently model-specific.
That choice avoids treating a model with a different prior as if it faced the
same conflicts, but it also prevents naive ranking by OR alone. Fixed
shared-row and consensus-hard analyses address denominator mismatch;
preservation measures anchor-induced corruption; and OR$\mid$Ret removes
matched rows with retrieval errors. The agreement across these analyses supports the dissociation while
leaving deployment-level rankings descriptive.

The broader implication is methodological. Long-context evaluations should pair
an access control with a task in which retrieved evidence must alter an already
plausible decision---for example, a distant state update paired with a locally
plausible action whose correct choice the update reverses, plus a direct state
query and a no-state control that preserve the local decision surface. This
applies beyond MWEs to updated user preferences, policy exceptions, tool state,
and document revisions. Conditional correction and preservation then distinguish
inaccessible state, accessible but behaviorally inert state, and harmful
overwriting; these conditional metrics should accompany overall task
accuracy, since
accuracy alone cannot separate correction from preservation of an
already-correct default. MWE-ECL offers one controlled instantiation of this
principle, not a claim that MWE errors directly predict agent failures or a
complete measure of long-context reasoning.

\section{Limitations}

The separate headline calls support a behavioral access--use dissociation but
do not reveal the interpretation call's hidden retrieval state; the three-model same-call
control omits the prior head.  Forced choice measures decisions rather than
mechanisms. Moreover, the no-anchor note and semantic anchor differ in content
and length, so their contrast is a benchmark-specific behavioral effect rather
than a semantics-only intervention on an internal prior.

The diagnostic intentionally uses an idiom-heavy set of strong-default MWEs, neutral
administrative filler, and only 25 English and 10 Chinese families. Chinese
difficulty is not matched to English, and the larger expansion lacks matched
human validation. Six headline panels are single-pass API snapshots;
coverage reflects API availability during the run window, and reasoning modes vary by endpoint.
Cross-grid rankings are descriptive, and
\originalcorrection{maximum-endpoint ECL values are right-censored tested-range
summaries rather than estimates of a monotone failure boundary.} Future work should broaden languages, document genres, response formats,
and deployments, and test whether extended reasoning closes the override gap.

\section{Conclusion}

MWE-ECL separates distant-anchor recoverability,
no-anchor defaults, and anchor-conditioned interpretation. On prior-conflict
rows, near-ceiling retrieval can coexist with substantially lower contextual
override. Targeted same-call and semantic-interference controls indicate that this
pattern is not explained solely by invocation separation or simple anchor loss.
These findings are behavioral and deployment-specific, but they support a
broader evaluation principle: long-context benchmarks should test not only
whether distant evidence can be recovered, but whether it changes a competing
local decision.

\bibliography{references}

\clearpage
\appendix
\raggedbottom
\makeatletter
\setlength{\@fptop}{0pt}
\setlength{\@dblfptop}{0pt}
\makeatother
\setcounter{table}{0}
\renewcommand{\thetable}{S\arabic{table}}
\providecommand{\mwepolish}[1]{#1}
\providecommand{\mwechange}[1]{#1}
\providecommand{\mwevfive}[1]{#1}

\section{Scope and Reading Guide}

This document supplies the evidence needed to audit the paper's behavioral
claims without reproducing internal run logs. It focuses on five questions:
(1) whether retrieval and override differ on identical conflict rows,
(2) whether the result depends on model-specific conflict sets,
(3) whether same-call and hard-cue controls preserve the pattern,
(4) whether repeated runs and the expanded catalog support robustness, and
(5) whether item labels and prompt construction pass independent checks.
The separate research code-and-data archive contains canonical row-level
outputs, analysis scripts, checksums, and a paper-to-file map. That archive
is not included in this arXiv submission.

The primary unit is a matched case--distance identity. Accuracy uses all valid
task-specific rows; override conditions on cases where the no-anchor prior is
wrong; OR$\mid$Ret additionally requires correct matched retrieval. Confidence
intervals use 10,000 whole-family bootstrap resamples. Exact within-model tests
apply sign flips to family-level paired differences and use Holm correction
over the eight prespecified headline panels.

\section{Evaluation Snapshot}

All prompts request one A/B label with temperature zero and a four-token visible
answer budget. The English headline grid contains 50 cases, 25 MWE families,
three task heads, and seven distances from 0 to 128K (1,050 records per model).
The API-interface panel is a time-stamped deployment comparison rather than an
architecture ranking. Exact request records and completed outputs are in the
code-and-data package, together with endpoint identifiers and access dates;
superseded retries and cost logs are excluded. Separate prompt-fit, hard-cue,
expansion, and Chinese regimes are not pooled with this grid.

\begin{table*}[t]
\centering
\footnotesize
\setlength{\tabcolsep}{4pt}
\begin{tabular}{llllr}
\toprule
Paper label & Reported endpoint identifier & Access dates & Grid & Rows \\
\midrule
DeepSeek V4 Flash & \texttt{deepseek-v4-flash} & 2026-05-16--07-15 & 0--128K & 1,050 \\
DeepSeek V4 Pro & \texttt{deepseek-v4-pro} & 2026-05-16--07-15 & 0--128K & 1,050 \\
Qwen3.6-Plus & \texttt{qwen3.6-plus} & 2026-05-16--07-15 & 0--128K & 1,050 \\
MiniMax-M3 & \texttt{MiniMax-M3} & 2026-07-10--07-15 & 0--128K & 1,050 \\
GLM-4.7 & \texttt{glm-4.7} & 2026-07-10--07-15 & 0--128K & 1,050 \\
GLM-5.1 & \texttt{glm-5.1} & 2026-05-23--07-16 & 0--128K & 1,050 \\
Qwen3.6-35B-A3B & \texttt{Qwen/Qwen3.6-35B-A3B} & 2026-07-19 & 0--128K & 1,050 \\
Gemma-4-26B-A4B-it & \texttt{google/gemma-4-26B-A4B-it} & 2026-07-19 & 0--128K & 1,050 \\
\bottomrule
\end{tabular}
\caption{Headline English deployment snapshots. Configuration-mismatched
stress regimes are reported separately and are not pooled with this grid.}
\label{tab:endpoints}
\end{table*}

\FloatBarrier
\section{Primary Statistical Accounting}

Table~\ref{tab:full-primary} gives the interval accounting omitted from the
compact main table. Every joint-ECL entry passes at the largest tested endpoint
under $\tau=0.8$ and is therefore a right-censored tested-range summary, not a
monotone failure-boundary estimate.

The canonical headline file contains exactly 8,400 rows: each of the eight
panels contributes 50 cases, three task heads, and seven distances. All planned
rows returned a parseable A/B label; there are no request errors or post-hoc
row exclusions in this grid. Option order is counterbalanced within each
case--distance cell, and all three heads are joined by the same case, distance,
and option-order identity before a conditional metric is computed.

Override uses a model-specific denominator because a conflict row is defined by
that model's own no-anchor prediction. Table~\ref{tab:conflict-consensus}
therefore reports both pairwise conflict-set overlap and a fixed 182-identity
consensus-hard subset. The former shows how similarly models expose their local
defaults; the latter supplies a common-denominator interpretation check. Neither
substitutes for override: they are cross-model sensitivity analyses for a metric
whose primary scientific target is within-model correction.

\begin{table*}[t]
\centering
\scriptsize
\setlength{\tabcolsep}{3pt}
\begin{tabular}{lcccc}
\toprule
Model & Interpretation [95\% CI] & Retrieval [95\% CI] & Override [95\% CI] ($n$) & Joint ECL \\
\midrule
Qwen3.6-Plus & 1.000 [1.000, 1.000] & 1.000 [1.000, 1.000] & 1.000 [1.000, 1.000] (174) & $\geq$128K \\
DeepSeek V4 Flash & 0.969 [0.926, 0.997] & 1.000 [1.000, 1.000] & 0.935 [0.851, 0.994] (170) & $\geq$128K \\
DeepSeek V4 Pro & 0.891 [0.837, 0.940] & 0.991 [0.983, 1.000] & 0.806 [0.695, 0.899] (175) & $\geq$128K \\
MiniMax-M3 & 0.949 [0.920, 0.974] & 0.991 [0.983, 1.000] & 0.911 [0.852, 0.963] (180) & $\geq$128K \\
GLM-4.7 & 1.000 [1.000, 1.000] & 1.000 [1.000, 1.000] & 1.000 [1.000, 1.000] (171) & $\geq$128K \\
GLM-5.1 & 1.000 [1.000, 1.000] & 1.000 [1.000, 1.000] & 1.000 [1.000, 1.000] (174) & $\geq$128K \\
Qwen3.6-35B-A3B & 0.931 [0.883, 0.974] & 0.997 [0.991, 1.000] & 0.870 [0.778, 0.949] (177) & $\geq$128K \\
Gemma-4-26B-A4B-it & 0.963 [0.926, 0.991] & 0.997 [0.991, 1.000] & 0.927 [0.851, 0.983] (178) & $\geq$128K \\
\bottomrule
\end{tabular}
\caption{Full English headline-grid accounting. Intervals use whole-family
bootstrap resampling; override is computed on each model's prior-conflict set.}
\label{tab:full-primary}
\end{table*}

\begin{table*}[t]
\centering
\scriptsize
\setlength{\tabcolsep}{3pt}
\begin{tabular}{lrrrrrrrr}
\toprule
Model & Conflict $n$ & Ret.C & OR & Gap & R$+$/I$-$ & I$+$/R$-$ & $p$ & Holm $p$ \\
\midrule
\WithinModelDissociationRows
\bottomrule
\end{tabular}
\caption{Within-model retrieval-control versus override tests on identical
prior-conflict rows. Discordances count matched retrieval-correct/interpretation-
wrong and interpretation-correct/retrieval-wrong events.}
\label{tab:within-model}
\end{table*}

\begin{table*}[t]
\centering
\scriptsize
\setlength{\tabcolsep}{4pt}
\begin{tabular}{lrrrrrr}
\toprule
Model & Pairs & Conflict & Override & Mean Jaccard & Consensus $n$ & Consensus acc. \\
\midrule
Qwen3.6-Plus & 350 & 174 & 1.000 & 0.820 & 182 & 1.000 \\
DeepSeek V4 Flash & 350 & 170 & 0.935 & 0.828 & 182 & 0.940 \\
DeepSeek V4 Pro & 350 & 175 & 0.806 & 0.734 & 182 & 0.813 \\
MiniMax-M3 & 350 & 180 & 0.911 & 0.807 & 182 & 0.918 \\
GLM-4.7 & 350 & 171 & 1.000 & 0.772 & 182 & 1.000 \\
Qwen3.6-35B-A3B & 350 & 177 & 0.870 & 0.813 & 182 & 0.890 \\
Gemma-4-26B-A4B-it & 350 & 178 & 0.927 & 0.818 & 182 & 0.929 \\
GLM-5.1 & 350 & 174 & 1.000 & 0.827 & 182 & 1.000 \\
\bottomrule
\end{tabular}
\caption{Conflict-set accounting on the shared 0--128K grid. Consensus
accuracy is fixed-subset interpretation accuracy, not a model-conditioned
override rate. Mean Jaccard summarizes pairwise conflict-set overlap.}
\label{tab:conflict-consensus}
\end{table*}

\begin{table*}[t]
\centering
\scriptsize
\begin{minipage}[t]{0.47\textwidth}
\centering
\begin{tabular}{lrrrr}
\toprule
Model & W$\rightarrow$C & W$\rightarrow$W & C$\rightarrow$C & C$\rightarrow$W \\
\midrule
\PrimaryTransitionRows
\bottomrule
\end{tabular}
\end{minipage}\hfill
\begin{minipage}[t]{0.51\textwidth}
\centering
\resizebox{\linewidth}{!}{%
\begin{tabular}{lrrrrr}
\toprule
Model & Pres. $n$ & Pres. [95\% CI] & Corrup. & Macro OR & Macro OR$\mid$Ret \\
\midrule
\PrimaryPreservationRows
\bottomrule
\end{tabular}}
\end{minipage}
\caption{Prior-to-interpretation transitions (left) and preservation with
family-macro sensitivity (right). W/C denote wrong/correct no-anchor priors.
Preservation conditions on prior-correct rows; corruption is its complement.}
\label{tab:transition-preservation}
\end{table*}

The transition counts make the headline pattern concrete. Across the five
non-ceiling panels, prior-conflict correction failures range from 11 to 34,
whereas prior-correct corruption ranges from zero to four. Interpretation loss
is therefore concentrated on rows that require changing an initially wrong
default, rather than on indiscriminate degradation after an anchor is added.
Family-macro override closely tracks the micro-pooled estimate, indicating that
the result is not driven by one unusually frequent MWE family.

\begin{table*}[t]
\centering
\scriptsize
\resizebox{\textwidth}{!}{%
\begin{tabular}{lrrrrrrrr}
\toprule
Model & Int.$_L$ & Int.$_F$ & $\Delta_{L-F}$ & OR$\mid$Ret$_L$ ($n$) & OR$\mid$Ret$_F$ ($n$) & Pres.$_L$ & Pres.$_F$ & Holm $p$ \\
\midrule
\SenseAsymmetryDetailedRows
\bottomrule
\end{tabular}}
\par\vspace{10pt}
\begin{minipage}[t]{0.52\textwidth}
\centering
\textit{No-anchor stability at 128K}\par\smallskip
\resizebox{\linewidth}{!}{%
\begin{tabular}{lrrrrr}
\toprule
Model & Dist. & Semantic agreement & Flip & Figurative & Option A \\
\midrule
\PriorStabilityEndpointRows
\bottomrule
\end{tabular}}
\end{minipage}\hfill
\begin{minipage}[t]{0.46\textwidth}
\centering
\textit{Interpretation retention at 128K}\par\smallskip
\resizebox{\linewidth}{!}{%
\begin{tabular}{lrrrr}
\toprule
Model & Dist. & Eligible $n$ & Retention [95\% CI] & Failure \\
\midrule
\RetentionEndpointRows
\bottomrule
\end{tabular}}
\end{minipage}
\par\vspace{10pt}
\textit{Exact matched-grid rates by distance (Interpretation/Retrieval/Override)}\par\smallskip
\resizebox{\textwidth}{!}{%
\begin{tabular}{lrrrrrrr}
\toprule
Model & 0 & 512 & 2K & 8K & 32K & 64K & 128K \\
\midrule
Qwen3.6-Plus & 1.000/1.000/1.000 & 1.000/1.000/1.000 & 1.000/1.000/1.000 & 1.000/1.000/1.000 & 1.000/1.000/1.000 & 1.000/1.000/1.000 & 1.000/1.000/1.000 \\
DeepSeek V4 Flash & 1.000/1.000/1.000 & 0.980/1.000/0.960 & 0.980/1.000/0.957 & 0.960/1.000/0.917 & 0.940/1.000/0.880 & 0.980/1.000/0.960 & 0.940/1.000/0.870 \\
DeepSeek V4 Pro & 1.000/1.000/1.000 & 0.920/1.000/0.846 & 0.880/1.000/0.769 & 0.860/0.980/0.783 & 0.820/1.000/0.680 & 0.880/1.000/0.760 & 0.880/0.960/0.792 \\
MiniMax-M3 & 0.980/1.000/0.962 & 0.940/1.000/0.880 & 0.920/0.980/0.852 & 1.000/1.000/1.000 & 0.980/1.000/0.960 & 0.960/1.000/0.923 & 0.860/0.960/0.800 \\
GLM-4.7 & 1.000/1.000/1.000 & 1.000/1.000/1.000 & 1.000/1.000/1.000 & 1.000/1.000/1.000 & 1.000/1.000/1.000 & 1.000/1.000/1.000 & 1.000/1.000/1.000 \\
GLM-5.1 & 1.000/1.000/1.000 & 1.000/1.000/1.000 & 1.000/1.000/1.000 & 1.000/1.000/1.000 & 1.000/1.000/1.000 & 1.000/1.000/1.000 & 1.000/1.000/1.000 \\
Qwen3.6-35B & 0.960/1.000/0.917 & 0.920/1.000/0.840 & 0.940/1.000/0.885 & 0.920/1.000/0.840 & 0.980/1.000/0.962 & 0.880/0.980/0.769 & 0.920/1.000/0.880 \\
Gemma-4-26B & 1.000/0.980/1.000 & 1.000/1.000/1.000 & 0.980/1.000/0.962 & 0.960/1.000/0.920 & 0.980/1.000/0.960 & 0.920/1.000/0.846 & 0.900/1.000/0.800 \\
\bottomrule
\end{tabular}}
\caption{Sense and endpoint diagnostics. Top: literal--figurative analysis;
subscripts L/F denote paired variants within each family, and Holm-adjusted
$p$ values come from exact two-sided sign-flip tests on family-level
interpretation differences. Bottom left: no-anchor semantic agreement with 0K,
disagreement (flip), semantic-figurative rate, and raw option-A rate. Bottom
right: interpretation retention among items correct at 0K and on matched 128K
retrieval; failure is one minus retention. Bottom: exact rates underlying the
distance curves, where override uses each model--distance cell's prior-conflict
denominator. Undefined family-level conflict denominators are omitted only from
the corresponding OR$\mid$Ret statistic.}
\label{tab:sense-endpoint-diagnostics}
\end{table*}

At 128K, no-anchor semantic choices agree with their 0K counterparts on
0.820--0.960 of cases, while raw option-A rates remain between 0.260 and 0.400.
This separates semantic default movement from a fixed answer-position bias.
The retention analysis asks a different question: among items already correct
at 0K and still retrieval-correct at 128K, how often does interpretation remain
correct? Retention ranges from 0.872 to 1.000, so endpoint failures are not
solely inherited from items that were already wrong at short context.

\FloatBarrier
\section{Same-Call and Semantic-Interference Controls}

The same-call control places retrieval and interpretation decisions after one
shared document and counterbalances query order. It removes invocation
separation but does not expose the prior head. At 64K and 128K, retrieval is
perfect in all DeepSeek V4 Pro and GLM-5.1 cells and in three of four MiniMax-M3
cells; interpretation remains below retrieval for DeepSeek V4 Pro and
MiniMax-M3 under both query orders. The hard-cue control instead changes
semantic interference while retaining a matched neutral condition.

The same-call grid contains 600 completed responses: three models, two
distances, two query orders, and 50 cases per cell. Both decisions are decoded
from one response, so the comparison cannot be attributed to separate model
invocations. Query order is counterbalanced because producing one label can
itself condition the second; the order-specific entries in
Table~\ref{tab:same-call} expose rather than average away this possibility.

\begin{table}[t]
\centering
\scriptsize
\setlength{\tabcolsep}{2.5pt}
\begin{tabular}{lcccc}
\toprule
& \multicolumn{2}{c}{64K} & \multicolumn{2}{c}{128K} \\
Model & Int.-first & Ret.-first & Int.-first & Ret.-first \\
\midrule
\SameCallDualHeadCompactRows
\bottomrule
\end{tabular}
\caption{Same-call control; entries are interpretation/retrieval accuracy over
50 English cases per cell.}
\label{tab:same-call}
\end{table}

Hard-cue records replace the neutral surroundings with either a target cue that
supports the anchor-licensed sense or a foil cue that reinforces the competing
local prior. Deltas are paired by family, case, distance, and task and are always
condition minus neutral. Thus a negative foil-interpretation delta denotes
greater interference, while a negative foil-prior delta means the foil moved
the no-anchor default in the same direction. Whole-family resampling keeps all
observations from an MWE family together.

\begin{table*}[t]
\centering
\scriptsize
\resizebox{\textwidth}{!}{%
\begin{tabular}{lrrrrrr}
\toprule
Model & Families & Paired $n$ & $\Delta$ target Int. [95\% CI] & $\Delta$ foil Int. [95\% CI] & $\Delta$ foil Prior [95\% CI] & $\Delta$ foil Ret. [95\% CI] \\
\midrule
\multicolumn{7}{l}{\textit{Matched 10-family pilot}} \\
\HardCuePairedPilotDeltaRows
\midrule
\multicolumn{7}{l}{\textit{25-family cue-triad runs}} \\
\HardCuePairedAllFamilyDeltaRows
\bottomrule
\end{tabular}}
\caption{Paired hard-cue changes relative to separately executed neutral
controls. Intervals use 10,000 whole-family bootstrap resamples. The pilot and
all-family panels differ in scope and are not pooled into a ranking.}
\label{tab:hard-cue}
\end{table*}

The paired cue deltas delimit two competing explanations. Target cues do not
systematically harm interpretation and improve it for DeepSeek V4 Pro in the
all-family run. Foil cues, by contrast, reduce interpretation for DeepSeek V4
Pro and Qwen3.6-Plus, while changing the no-anchor prior more strongly for most
panels. Retrieval remains unchanged for four of six all-family panels and
drops for DeepSeek V4 Pro and MiniMax-M3. Thus semantic interference can affect
both recoverability and decision use, but the two effects are not identical.
Because the pilot and all-family runs differ in catalog coverage and execution
window, the table supports within-panel cue contrasts rather than a model
ranking.

\FloatBarrier
\section{Repeatability and Position Pilot}

Three-pass endpoint checks quantify API-interface nondeterminism for the two
open-weight model endpoints. The 128K passes reproduce identical aggregate
accuracies and labels. The position probe is deliberately descriptive because
each cell contains only 20 matched items.

Each repeatability cell consists of three independently executed 50-record
passes. Agreement is the fraction of item identities receiving the same label
in all passes, whereas the reported standard deviation summarizes pass-level
accuracy. The small 64K Qwen variation therefore reflects a few label changes,
not missing records; all four 128K interpretation/retrieval cells have perfect
cross-pass label agreement. These checks quantify deployment variation and do
not turn a three-pass snapshot into a model-level stability guarantee.

\begin{table}[t]
\centering
\scriptsize
\setlength{\tabcolsep}{2.5pt}
\begin{tabular}{llrrr}
\toprule
Model & Task & Dist. & Mean $\pm$ std. & Agreement \\
\midrule
Qwen-35B & Retrieval & 64K  & $0.987\pm0.012$ & 0.980 \\
          & Interpretation & 64K  & $0.900\pm0.020$ & 0.960 \\
          & Prior & 64K  & $0.507\pm0.031$ & 0.920 \\
          & Retrieval & 128K & $1.000\pm0.000$ & 1.000 \\
          & Interpretation & 128K & $0.920\pm0.000$ & 1.000 \\
          & Prior & 128K & $0.500\pm0.000$ & 1.000 \\
\midrule
Gemma-26B & Retrieval & 64K  & $1.000\pm0.000$ & 1.000 \\
          & Interpretation & 64K  & $0.920\pm0.000$ & 1.000 \\
          & Prior & 64K  & $0.500\pm0.000$ & 1.000 \\
          & Retrieval & 128K & $1.000\pm0.000$ & 1.000 \\
          & Interpretation & 128K & $0.900\pm0.000$ & 1.000 \\
          & Prior & 128K & $0.500\pm0.000$ & 1.000 \\
\bottomrule
\end{tabular}
\caption{Endpoint repeated-run accuracy and per-item label agreement over
three 50-record passes per cell. Qwen-35B and Gemma-26B abbreviate the model
labels used in the main paper.}
\label{tab:repeatability}
\end{table}

\begin{table}[t]
\centering
\scriptsize
\setlength{\tabcolsep}{3pt}
\begin{tabular}{lrlrrrr}
\toprule
Model & Dist. & Pos. & $n$ & Ret. correct & IUR & Override \\
\midrule
\AnchorPositionRows
\bottomrule
\end{tabular}
\caption{Preliminary Qwen3.6-35B-A3B anchor-position probe. The small-$N$
design does not support a general position-effect claim; Ret. correct is a
count out of $n$, while IUR and override are rates.}
\label{tab:position}
\end{table}

At 128K, all four interpretation/retrieval repeatability cells have identical
labels across the three passes. The 64K Qwen cells show the only pass-level
variation, with agreement of 0.920--0.980 depending on the head. The position
pilot contains no monotone late-to-early ordering: at 8K, override ranges from
0.70 to 0.90 across positions, while at 64K it is 0.60--0.70. These 20-item
cells are useful as a failure-mode check but are too small to estimate a
general positional curve.

\section{Prompt-Fit Stress Beyond 128K}

Prompt-fit runs test whether the access--use pattern remains visible when a
deployment is evaluated near its practical serving limit. They use the same
three task heads and 50 English cases, but endpoint-specific filler sizes and
tokenizers make them descriptive extensions rather than additional matched-grid
distance cells. All 600 responses complete without provider errors or unparsed
labels.

\begin{table}[t]
\centering
\scriptsize
\setlength{\tabcolsep}{3pt}
\resizebox{\columnwidth}{!}{%
\begin{tabular}{llrrrr}
\toprule
Model & Context & Int. & Ret. & Prior & Prompt tokens \\
\midrule
Qwen3.6-35B & 256K & 0.940 & 1.000 & 0.480 & 254,910--254,952 \\
Gemma-4-26B & 256K & 0.880 & 1.000 & 0.500 & 245,217--245,256 \\
DeepSeek V4 Flash & 512K & 0.860 & 1.000 & 0.500 & 549,174--549,217 \\
DeepSeek V4 Flash & 1M & 0.880 & 1.000 & 0.480 & 1,030,454--1,030,497 \\
\bottomrule
\end{tabular}}
\caption{Prompt-fit stress outside the matched grid. Requested filler is
217,088 tokens for the 256K rows, 512K for the 512K row, and 960K for the 1M
row. Each row contains 150 records (50 cases across three heads), with no
request errors or unparsed labels. Exact physical prompt-token ranges are
retained for reproducibility.}
\label{tab:prompt-fit}
\end{table}

All four prompt-fit rows retain perfect explicit retrieval, whereas
interpretation ranges from 0.860 to 0.940. The DeepSeek V4 Flash 1M row is not
worse than its 512K row, reinforcing that these independently prompted cells
should not be read as a monotone degradation curve. Their role is narrower:
the access--use dissociation remains observable near endpoint-specific serving
limits when the complete prompt physically fits.

\FloatBarrier
\section{Dataset Validation and Catalog Expansion}

The English primary catalog is fixed before model evaluation. Two independent
annotators selected the anchor-licensed answer for all 50 items without seeing
model outputs; both matched all intended labels and raised no flags. Automated
checks verify schema, task/distance balance, option order, and prompt structure.
A static audit finds no verbatim answer-option leakage and no target-MWE leakage
from the neutral filler bank. Four literal anchors repeat the physical target
expression by design and are reported in the released audit.

\begin{table}[t]
\centering
\small
\setlength{\tabcolsep}{5pt}
\begin{tabular}{lrr}
\toprule
Audit & English & Chinese \\
\midrule
Cases reviewed & 50 & 20 \\
Independent annotators & 2 & 2 \\
Judgments matching intended labels & 100/100 & 40/40 \\
Flagged items & 0 & 0 \\
Verbatim option leakage & 0 & 0 \\
Target-MWE leakage from neutral filler & 0 & 0 \\
\bottomrule
\end{tabular}
\caption{Human label validation and static leakage checks. Annotators were
blind to model outputs; the two languages were reviewed separately.}
\label{tab:human-leakage-audit}
\end{table}

The audited expansion adds 60 English cases from 30 additional families and is
evaluated separately at 0/8K/64K. It broadens lexical and constructional
coverage but is not difficulty-matched to the primary suite and is not used for
headline rankings.

\begin{table}[t]
\centering
\scriptsize
\setlength{\tabcolsep}{3pt}
\begin{tabular}{lrrrrr}
\toprule
Model & Int. & Ret. & Override & Conflict & Primary OR \\
\midrule
DeepSeek V4 Pro & 0.944 & 0.961 & 0.897 & 87 & 0.806 \\
GLM-4.7 & 0.933 & 0.961 & 0.881 & 84 & 1.000 \\
GLM-5.1 & 0.922 & 0.983 & 0.878 & 90 & 1.000 \\
MiniMax-M3 & 0.928 & 0.983 & 0.867 & 90 & 0.911 \\
\bottomrule
\end{tabular}
\caption{Audited 60-case catalog-expansion reproduction. Each model has 540
records, zero request errors, and zero unparsed responses. Primary OR is shown
only for context; the two catalogs are not pooled.}
\label{tab:expansion}
\end{table}

A descriptive inventory-sensitivity split is included in the released derived
tables. Because the inventory-absent group is not a held-out set, it does not
support a contamination claim and is not used in the paper's conclusions.

\begin{table}[t]
\centering
\scriptsize
\setlength{\tabcolsep}{3pt}
\resizebox{\columnwidth}{!}{%
\begin{tabular}{lrrrrr}
\toprule
Model & Interpretation & Retrieval & Conflict $n$ & Override & Preservation \\
\midrule
\ChineseBehavioralRows
\bottomrule
\end{tabular}}
\caption{Aggregate results on the separately reported 20-case, 10-family
Chinese subset. The subset uses the same three heads and seven nominal
distances, but is neither a translation nor a difficulty-matched counterpart
of the English catalog.}
\label{tab:chinese-aggregate}
\end{table}

The Chinese subset reproduces the qualitative contrast for several panels:
DeepSeek V4 Pro, MiniMax-M3, Qwen3.6-35B, and Gemma-4-26B all have lower
override than retrieval, while three ceiling panels remain at 1.000. DeepSeek
V4 Flash is near ceiling on both measures. This is a multilingual robustness
check, not evidence that the two catalogs have equal difficulty: the Chinese
items are independently sourced, smaller, and reported only in aggregate.

\FloatBarrier
\section{Prompt and Release Boundary}

The system instruction is: ``Use the entire document, including distant
context, to choose the correct interpretation. Reply with exactly one letter:
A or B. Do not explain.'' The user prompt contains an anchor note, deterministic
neutral filler, a final event, the task question, and two options. Retrieval
replaces the interpretation question with a direct anchor-scenario query;
prior-control replaces the anchor with a note stating that no disambiguating
background is provided. Complete prompt records are included in the data
package rather than duplicated here.

The Chinese subset is reported only in aggregate because its source inventory
uses controlled release. It contains 20 cases across 10 families, is not a
difficulty-matched translation of the English suite, and is analyzed separately.
Two native-Chinese annotators matched all intended labels with no flags. The
code-and-data package includes the transformation and validation code, but not
the controlled inventory or row-level Chinese outputs.

\section{Interpretive Boundary of the Controls}

The controls address distinct alternatives and should not be combined into a
single mechanistic claim. Same-call evaluation rules out invocation separation
for its three tested deployments, but does not reveal the latent state used to
produce either answer. Hard-cue deltas establish sensitivity to structured
semantic interference, but a retrieval decrease in some panels prevents a pure
integration-only interpretation. Repeated runs constrain deployment variation
at selected endpoints, whereas the position pilot and prompt-fit rows remain
descriptive because their grids are deliberately smaller or configuration
specific.

Taken together, these checks support a behavioral conclusion: an explicit
matched retrieval answer can be correct while the associated interpretation
still follows the competing local default. They do not identify a neural
mechanism, prove that every interpretation error contains a recoverable hidden
anchor representation, or establish a universal context-length ranking.

\section{Residual Limitations}

The benchmark is an idiom-heavy forced-choice diagnostic, not an estimate of
average language understanding. Conflict sets are model-specific, deployment
snapshots can drift, Chinese and English difficulty are not matched, and the
expansion is not held out. Separate-call retrieval controls do not reveal the
interpretation call's hidden state; the same-call control addresses invocation
separation for three models but omits the prior head. These constraints bound
the claims to behavioral access and use under the reported deployments.

\end{document}